\documentclass{IEEEoj}
\usepackage[style=ieee, sorting=none]{biblatex}
\usepackage{booktabs}
\usepackage{tabularx}

\usepackage{float}

\usepackage{hyperref}

\usepackage{amsmath,amssymb,amsfonts}
\usepackage{algorithmic}
\usepackage{graphicx,color}
\usepackage{textcomp}
\def\BibTeX{{\rm B\kern-.05em{\sc i\kern-.025em b}\kern-.08em
    T\kern-.1667em\lower.7ex\hbox{E}\kern-.125emX}}
\AtBeginDocument{\definecolor{ojcolor}{cmyk}{0.93,0.59,0.15,0.02}}
\def\OJlogo{\vspace{-14pt}\includegraphics[height=28pt]{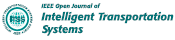}}
\begin{document}
\receiveddate{XX Month, XXXX}
\reviseddate{XX Month, XXXX}
\accepteddate{XX Month, XXXX}
\publisheddate{XX Month, XXXX}
\currentdate{XX Month, XXXX}
\doiinfo{OJITS.2022.1234567}


\title{UrbanTwin: Synthetic Roadside LiDAR Datasets}
\author{MUHAMMAD SHAHBAZ\authorrefmark{1}
, SHAURYA AGARWAL\authorrefmark{2}
}
\affil{University of Central Florida, 4000 Central Florida Blvd, Orlando, FL 32816 USA}
\corresp{CORRESPONDING AUTHOR: Muhammad Shahbaz (e-mail: muhammad.shahbaz@ucf.edu).}
\markboth{Preparation of Papers for IEEE OPEN JOURNALS}{Author \textit{et al.}}

\begin{abstract}
This article presents \textbf{\texttt{UrbanTwin}} datasets—high-fidelity, realistic replicas of three public roadside lidar datasets: \textbf{\texttt{LUMPI}}, \textbf{\texttt{V2X-Real-IC}}, and \textbf{\texttt{TUMTraf-I}}. Each \texttt{UrbanTwin} dataset contains $10K$ annotated frames corresponding to one of the public datasets. Annotations include 3D bounding boxes, instance segmentation labels, and tracking IDs for six object classes, along with semantic segmentation labels for nine classes.
These datasets are synthesized using emulated lidar sensors within realistic digital twins, modeled based on surrounding geometry, road alignment at lane level, and the lane topology and vehicle movement patterns at intersections of the actual locations corresponding to each real dataset. Due to the precise digital twin modeling, the synthetic datasets are well aligned with their real counterparts, offering strong standalone and augmentative value for training deep learning models on tasks such as 3D object detection, tracking, and semantic and instance segmentation.
We evaluate the alignment of the synthetic replicas through statistical and structural similarity analysis with real data, and further demonstrate their utility by training 3D object detection models solely on synthetic data and testing them on real, unseen data. 
The high similarity scores and improved detection performance, compared to the models trained on real data, indicate that the \texttt{UrbanTwin} datasets effectively enhance existing benchmark datasets by increasing sample size and scene diversity. In addition, the digital twins can be adapted to test custom scenarios by modifying the design and dynamics of the simulations. To our knowledge, these are the first digitally synthesized datasets that can replace \textit{in-domain} real-world datasets for lidar perception tasks. \texttt{UrbanTwin} datasets are publicly available at \url{https://dataverse.harvard.edu/dataverse/ucf-ut}.
\end{abstract}



\maketitle

\section{INTRODUCTION}

\textbf{{M}{otivation and Context}}: The precision and robustness of light detection and ranging (lidar) technology are becoming foundational for the advancement of perception algorithms for intelligent transportation systems (ITS). In this context, high-quality roadside lidar datasets are essential, particularly for training and evaluation of 3D perception algorithms in infrastructure-assisted vehicle environments. Although real-world datasets such as LUMPI \cite{busch2022lumpi}, V2X-Real \cite{xiang2024v2x}, TUMTraf-I \cite{zimmer2023tumtraf}, and others \cite{wang2022ips300+, mirlach2025r, yu2022dair, yongqiang2021baai, ye2022rope3d} provide valuable benchmarks, expanding them remains a challenging task that requires substantial human effort, time, and financial resources.

\begin{figure}[t]
  \centering
  \begin{minipage}[t]{0.48\textwidth}
    \includegraphics[width=\linewidth]{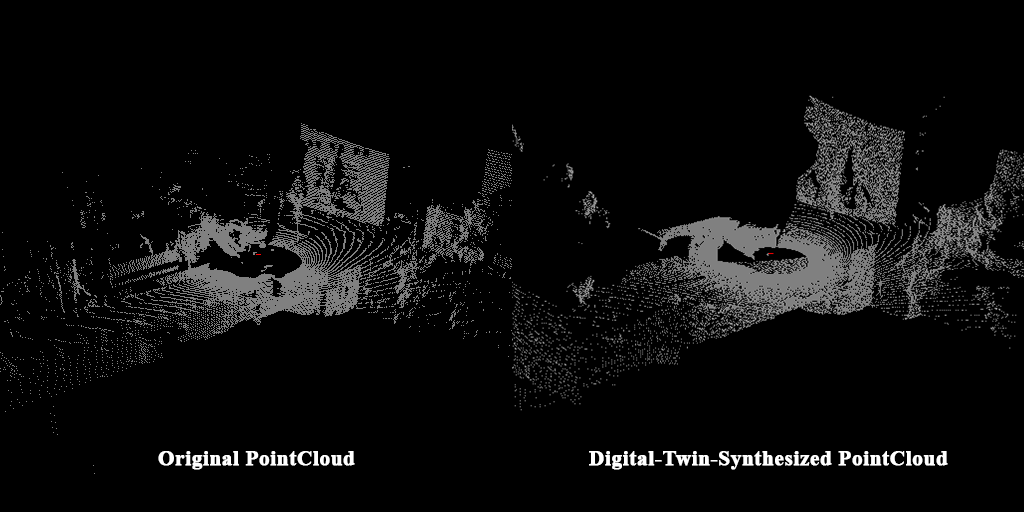}
    \caption{Lidar data synthesis from realistic digital twins yields close-to-real data. \textbf{Left}: A point cloud frame from real-world V2X-Real \cite{xiang2024v2x} dataset. \textbf{Right}: A digital-twin-synthesized point cloud for the same dataset. Can you notice a high similarity?}
    \label{fig:orig_vs_dt_synth}
  \end{minipage}
\end{figure}

\begin{figure*}[h]
\centerline{\includegraphics[width=0.98\textwidth]{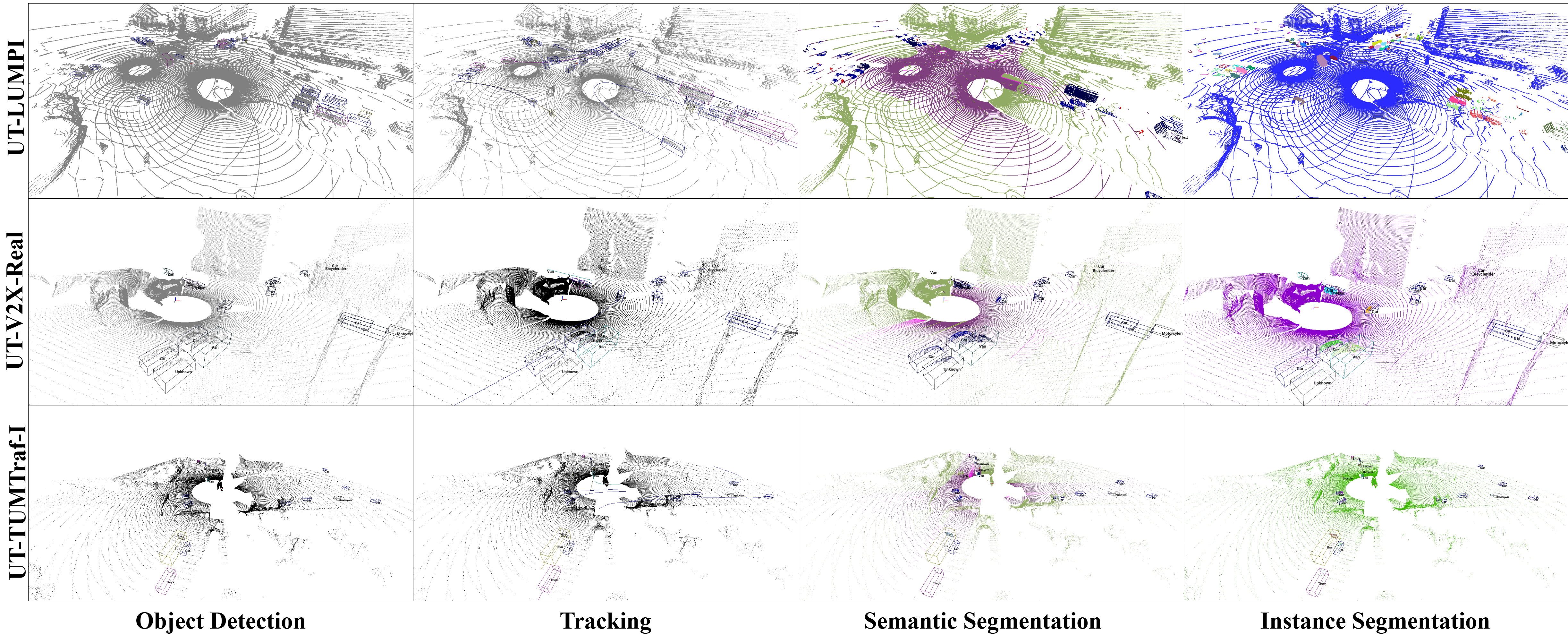}}
\caption{\texttt{UrbanTwin} datasets support all four major tasks for lidar-based perception. They provide 3D bounding boxes for object detection, object IDs for tracking, and KITTI \cite{geiger2013vision} style labels for both semantic-level and instance-level segmentation. Each frame includes up to 6 object classes and up to 9 semantic-level categories.
\label{fig:supported_tasks}}
\end{figure*}

\noindent \textbf{Limitations of Current Datasets and Simulators}: Simulation environments offer a scalable approach to generate lidar datasets \cite{li2024choose}. Comprehensive open-source simulators such as CARLA \cite{dosovitskiy2017carla}, DeepDrive \cite{team2020deepdrive}, and Vista \cite{amini2022vista} support a wide range of tasks in perception, planning, and control for autonomous driving systems (ADS). This functionality can often be extended to support perception tasks for roadside sensors as well. However, these simulators typically rely on hand-crafted 3D assets and simplified physics assumptions, resulting in simulations that lack alignment with real-world environments and introduce a significant sim-to-real domain gap \cite{manivasagam2020lidarsim}. As a result, models trained on synthetic data from such simulations often exhibit poor performance when deployed in real-world settings \cite{8864642}.

\noindent \textbf{Digital Twins for Dataset Synthesis}: To bridge the gap between synthetic and real data, it is essential to develop simulations that incorporate both the physical structure and behavioral dynamics of real-world environments. In this paper, we present synthetic datasets  constructed from high-fidelity digital twins of real-world scenes, based on three established roadside lidar benchmarks: (a) the Leibniz University Multi-perspective Intersection (LUMPI) dataset \cite{busch2022lumpi}, (b) the Large-scale Dataset for Vehicle-to-Everything Cooperative Perception (V2X-Real) \cite{xiang2024v2x}, and (c) the TumTraf Intersection (TUMTraf-I) dataset \cite{zimmer2023tumtraf}. Our digital twins are carefully modeled to replicate not only physical aspects such as geographic layout and road geometry, but also behavioral characteristics, including sensor specifications and typical traffic maneuver patterns. To enhance the utility of the synthetic datasets, dynamic elements in the simulations, such as the types of road users (cars, trucks, etc.) and traffic flow, are modeled stochastically. This enables the generation of unique yet realistically aligned datasets corresponding to their real-world counterparts.

\noindent \textbf{Empirical Validation of Synthetic Data}: We demonstrate that by leveraging high-fidelity structural modeling of real locations using accurate 3D geometry information, including lane-level road features (e.g., height, superelevation), buildings, and vegetation, our digital twins offer a more realistic static representation of the physical environment. Additionally, by incorporating high-level dynamic elements such as typical vehicle maneuver patterns at intersections and matching the sensor specifications of the real roadside lidars, the synthetic datasets achieve stronger alignment with real-world conditions (see Fig.~\ref{fig:orig_vs_dt_synth}). Finally, since these datasets are generated through simulation, they inherently support all four core perception tasks: 3D object detection, tracking, semantic segmentation, and instance segmentation (see Fig.~\ref{fig:supported_tasks}).

The digital-twin-synthesized replicas of established roadside lidar datasets exhibit minimal domain gap. Unlike many synthetic datasets used in ITS research, these datasets are purpose-built to closely align with their real-world counterparts, making them valuable tools for augmenting and enhancing existing benchmarks. We apply distribution analysis techniques to compare the synthetic datasets with real data.

To further empirically validate the utility of these synthesized replicas, we focus on the 3D object detection task using the LUMPI \cite{busch2022lumpi} and V2X-Real \cite{xiang2024v2x} datasets. For this evaluation, two established off-the-shelf models, SEED \cite{liu2024seed} and SECOND \cite{yan2018second}, are employed. In the LUMPI case, SEED is trained exclusively on digital-twin-synthesized data and compared against another SEED model trained from scratch, under identical settings, on the training subset of the real dataset. Both models are evaluated on the real test subset using the KITTI \cite{geiger2013vision} benchmark.

For the V2X-Real case, SECOND is trained solely on the synthetic dataset and evaluated on the real test set of V2X-Real-IC. The results are compared against benchmark models reported in the original V2X-Real paper. The findings show that models trained on synthetic data perform on par with, and in some cases outperform, models trained on real data, setting a new standard for the utility of synthetic datasets in roadside lidar perception tasks.

\newpage
To summarize, the key \textbf{contributions} of this paper are as follows:
\begin{itemize}
\item To the best of our knowledge, this is the first effort to synthesize lidar datasets specifically designed to augment established real-world roadside benchmarks.
\item We propose a novel use of realistic digital-twin modeling that integrates both static elements (e.g., geometry and lane-level alignment) and dynamic behaviors (e.g., traffic maneuver patterns) of real-world scenes for accurate data synthesis.
\item We demonstrate that the synthesized dataset replicas are closely aligned with their real counterparts through extensive statistical and structural similarity analyses.
\item This is the first study to show that models trained entirely on simulation data can match or exceed the performance of those trained on real data for lidar-based 3D object detection tasks.
\end{itemize}

It is important to note that this work focuses exclusively on the Lidar modality to rigorously evaluate the quality of synthetic point cloud data. While camera data is a valuable component in multi-modal perception systems, it is intentionally excluded from the scope of this study to ensure a controlled analysis of Lidar-centric, sim-to-real transfer.

\section{Related Work}
\textbf{Real-world Roadside Lidar Datasets:}
Recent research has increasingly focused on using lidar technology for sensing road users from an infrastructure perspective, leading to the development of several roadside lidar datasets. In 2021, the BAAI-VANJEE dataset \cite{yongqiang2021baai} was introduced, capturing highway and intersection scenarios across various times of day and weather conditions. In 2022, the IPS300+ dataset \cite{wang2022ips300+} was released, notable for featuring over 300 road users per frame, primarily due to the high volume of pedestrians and cyclists at a major intersection in Beijing. That same year, the DAIR-V2X dataset \cite{yu2022dair} became the first large-scale, multi-view dataset designed for vehicle-infrastructure cooperative driving research. Also in 2022, LUMPI \cite{busch2022lumpi} was introduced as the largest dataset by frame count, collected over multiple days and weather conditions. Rope3D \cite{ye2022rope3d}, while primarily focused on monocular 3D object detection using cameras, incorporated lidar data to improve annotation accuracy. In 2023, the TUMTraf dataset \cite{zimmer2023tumtraf}—a successor to the A9 dataset \cite{cress2022a9}—was released, with a focus on non-pedestrian traffic. The V2X-Real dataset \cite{xiang2024v2x}, introduced in 2024, was the first large-scale dataset to integrate multiple vehicles and smart infrastructure for V2X cooperative perception. This was complemented by other large-scale efforts like RCooper \cite{hao2024rcooper}, which provides a real-world dataset focused on roadside cooperative perception. Concurrently, advancements in spatio-temporal fusion for multi-agent perception and prediction were explored in works like V2xpnp \cite{zhou2024v2xpnp}. That same year, Int2Sec \cite{tang2024multi} targeted urban intersections using a digital twin perspective, offering 10 scenes annotated via dual roadside lidars. HoloVIC \cite{ma2024holovic} further expanded the space by providing over 100K synchronized frames with 3D object annotations across multiple holographic intersections. Most recently, in 2025, the R-LiViT dataset \cite{mirlach2025r} was introduced, combining lidar, RGB, and thermal imaging from a roadside viewpoint, with a focus on pedestrians and other vulnerable road users.

Despite this growing body of work, creating real-world labeled lidar datasets remains a challenging and resource-intensive task. It demands substantial human effort, time, and financial investment. The complexity of object shapes and motions, frequent occlusions, and varying backgrounds in point cloud data, particularly from the roadside perspective, make the annotation process labor-intensive and highly dependent on skilled expertise.

\noindent\textbf{Synthetic Roadside Lidar Datasets:} 
The recent surge in roadside lidar datasets reflects a growing interest in infrastructure-centric perception. However, creating large-scale, annotated lidar datasets remains an inherently expensive and labor-intensive process, restricting the scalability needed for training safety-critical models \cite{kalra2016driving}. This challenge has accelerated the adoption of simulation, which can speed up research by orders of magnitude \cite{feng2023dense}, and is increasingly motivated by the Digital Twin (DT) paradigm. Foundational work has established how infrastructure-based digital twins can serve as quality-controlled information sources for automated driving functions \cite{OJITS_DT_2022}. As detailed in recent surveys, the DT-ITS framework aims to create virtual replicas of entire transportation ecosystems for full lifecycle management and the validation of perception algorithms \cite{Ge2024DTITS, Gu2025DigitalTwin}. In response to these needs, and to overcome the limitations of real-world data collection, several synthetic lidar datasets have been proposed.

In 2022, SynLidar \cite{xiao2022transfer} used a custom simulator to synthesize a large synthetic lidar dataset generated using multiple simulated scenes. In the same year, V2X-Sim and V2X-Set \cite{li2022v2x} datasets simulated comprehensive multi-agent scenarios to generate synchronized camera and lidar data that included roadside units (RSUs) into the mix for cooperative perception tasks. DOLPHINS \cite{mao2022dolphins} dataset included six autonomous driving scenarios involving temporarily synchronized sensor data from an ego vehicle. In 2023, the DeepAccident dataset \cite{wang2024deepaccident} was released, providing 57K frames that include safety-critical scenarios for evaluating accident prediction models. In 2024, SynthmanticLiDAR \cite{montalvo2024synthmanticlidar} introduced a CARLA-based simulator tailored for lidar semantic segmentation, producing annotated point clouds aligned with real-world class distributions. TUMTraf Synthetic (TUMTraf-Syn) Dataset \cite{zhou2024warm} included 24,000 images with depth maps, and 2D and 3D annotations in 10 object categories focusing sim2real monocular 3D object detection. The SCOPE dataset \cite{gamerdinger2024scope}, also introduced in 2024, emphasized diversity, featuring realistic sensor models, physically accurate weather conditions, a catalog of over 40 scenarios, up to 24 collaborative agents, and passive traffic.

It is important to note, however, that these synthetic lidar datasets primarily focus on vehicle-infrastructure cooperative (VIC) driving scenarios. To the best of our knowledge, no synthetic lidar dataset to date has been developed specifically for standalone roadside lidar applications.

\noindent\textbf{Sim-to-Real Gap Mitigation Approaches}: Synthetic datasets are often created using simulators that model 3D scenes under user-defined conditions. However, the outputs frequently diverge from real data due to unmodeled environmental and sensor-specific effects, resulting in a noticeable sim-to-real domain gap. For Lidar-based 3D object detection, this gap can be particularly severe; recent studies have quantified this effect, showing that state-of-the-art detectors trained exclusively on simulated data can suffer a performance degradation of over 50\% in mean Average Precision (mAP) when evaluated on real-world data \cite{Michele2023Sim2Real}. Addressing this gap has been a central focus in many recent studies.

Manivasagam et al. \cite{Manivasagam2023ZeroGap} developed a paired-scenario methodology for evaluating domain discrepancies by digitally reconstructing real scenes, enabling direct comparisons between real and simulated lidar data under identical conditions. Their work emphasizes the need to model several physics-based factors, such as multi-echo pulses, motion distortion, and material reflectance, to achieve realism. Similarly, Haider et al. \cite{Haider2022HighFidelity} proposed a lidar model incorporating accurate ray-tracing and a complete signal-processing pipeline, validated against real-world measurements. Their results show that sensor imperfections—such as optical losses, electronic noise, and multi-echo behavior—must be accurately modeled, as neglecting these effects significantly reduces simulation fidelity.

In parallel, several data-driven approaches have aimed to improve realism. For example, Haghighi et al. \cite{Haghighi2024CoLiGen} introduced CoLiGen, a generative framework that converts lidar data to depth-reflectance images and uses GANs to translate simulated scans into more realistic point clouds. Domain adaptation methods have also been applied to reduce distributional mismatch. Xiao et al. \cite{xiao2022transfer} (SynLiDAR) proposed disentangling point cloud appearance and density differences, leveraging a GAN-based translator to align synthetic data with real-world distributions. More recently, Saltori et al. \cite{Saltori2024Mix} presented compositional semantic mixing—an unsupervised approach that combines semantic components from synthetic and real-world point clouds using a dual-branch network, significantly enhancing segmentation performance on target data.

Most of these approaches mitigate the sim-to-real gap by applying data-driven techniques after the dataset has already been generated. However, recent studies emphasize the importance of incorporating detailed 3D assets and sensor models during the simulation process itself. Despite this, there remains a lack of synthetic datasets that offer high-fidelity geometry, accurate sensor emulation, and realistic motion dynamics—all critical components for minimizing the domain gap at the source.

\noindent\textbf{Gaps in Existing Work \& Motivation}: While recent studies and datasets have significantly advanced roadside lidar perception research, several critical gaps remain unaddressed: (a) existing synthetic datasets predominantly target vehicle-infrastructure cooperative (VIC) scenarios, neglecting datasets solely dedicated to roadside lidar applications, (b) existing approaches primarily address sim-to-real gaps through post-hoc, data-driven adaptations rather than foundational improvements in simulation realism itself, and (c) creating real-world labeled lidar datasets remains inherently expensive, labor-intensive, and time-consuming, often requiring substantial human expertise. 

To date, synthetic datasets leveraging accurate geometric modeling, detailed sensor characteristics, and realistic motion dynamics are lacking. Motivated by these limitations, this article presents the \texttt{UrbanTwin} datasets, a class of high-fidelity synthetic replicas of real-world roadside lidar datasets. By employing detailed digital-twin modeling that encapsulates both static geometry and dynamic traffic behaviors of real-world environments, our aim is to bridge existing sim-to-real gaps. The resulting datasets not only enhance existing benchmarks but also offer a viable alternative to real-world data for critical lidar-based perception tasks.

\section{UrbanTwin Dataset Generation and Description}
We present three synthetic roadside lidar datasets: \texttt{UrbanTwin-LUMPI} \cite{ut_lumpi}, \texttt{UrbanTwin-V2X-Real} \cite{ut_v2x_real}, and \texttt{UrbanTwin-TUMTraf-I} \cite{ut_tumtraf_i}. For brevity, these datasets will be referred to using the \texttt{UT-} prefix throughout this article. Each synthetic dataset is designed to replicate the core characteristics of its corresponding real-world counterpart, LUMPI \cite{busch2022lumpi}, V2X-Real \cite{xiang2024v2x}, and TUMTraf-I \cite{zimmer2023tumtraf}.

\begin{figure*}[h]
\centerline{\includegraphics[width=0.98\textwidth]{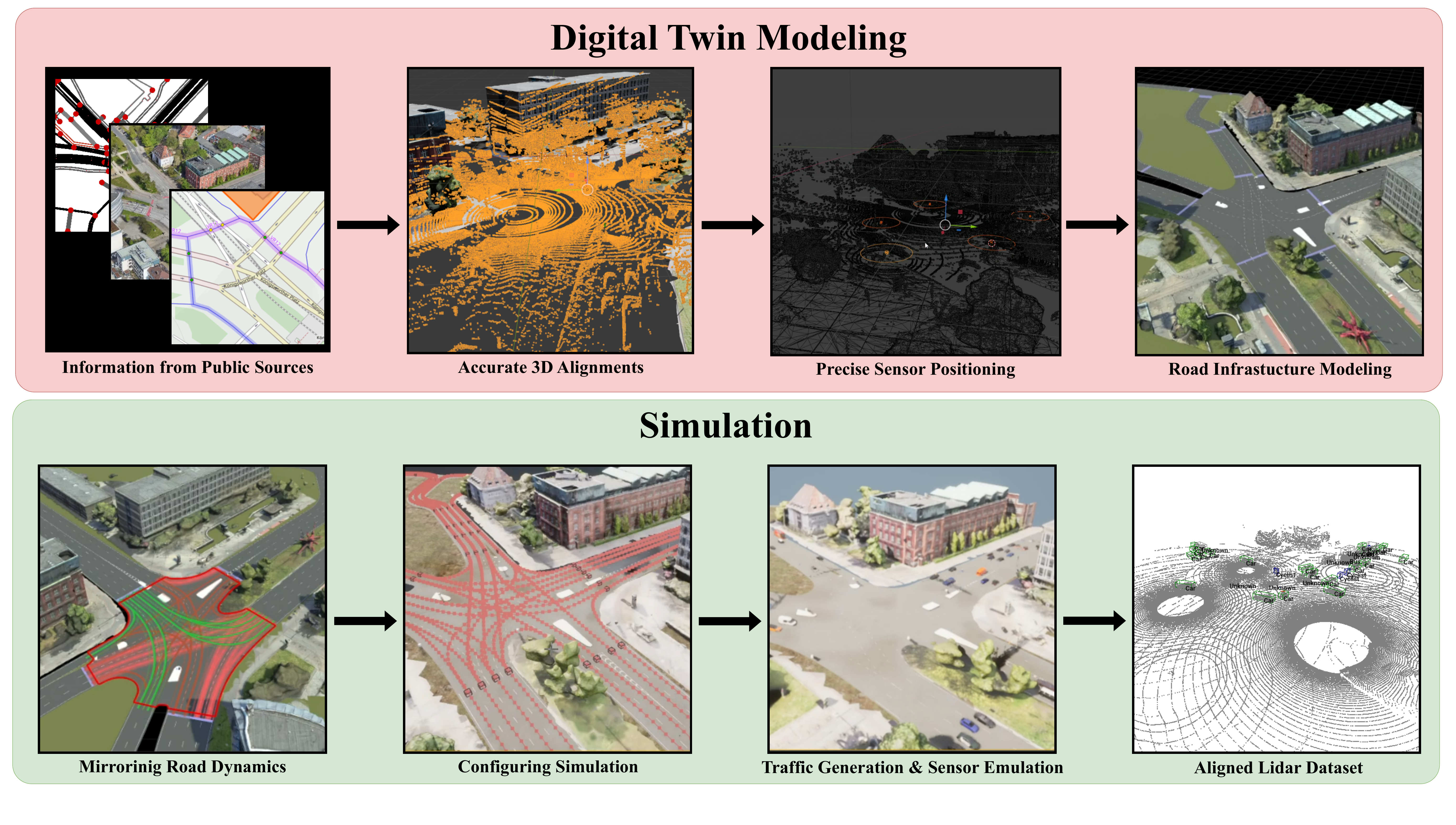}}
\caption{Overview of Synthetic Dataset Generation. Top: First, a 3D model of a real-world scene is created utilizing publicly available information, including road network information from OpenStreetMaps \cite{OpenStreetMap}, satellite imagery, etc. Then, the model is fine-aligned to the real point cloud frame, followed by embedding positional information of the lidar sensors. Finally, the road is constructed. Bottom: Simulation is modeled using real-world dynamics of the traffic, followed by importing all the assets to a custom CARLA map, and traffic is generated stochastically but conforms statistically to the target real dataset. Finally, the point cloud data along with label information is stored.
\label{fig:data_gen_process}}
\end{figure*}

\subsection{Generation Approach} 
The datasets are generated using the CARLA simulator \cite{dosovitskiy2017carla} by running simulations on custom maps constructed as digital twins. These maps are built from publicly available geographic and structural data. Key elements—including surrounding buildings, road geometry (e.g., prominent vegetation, raised medians), intersection layout and dynamics (e.g., vehicle maneuvers and signalization), and sensor placement and specifications (e.g., position, tilt, angular resolution, field of view)—are carefully modeled to ensure spatial and sensory alignment with the original datasets. A high-level overview of the dataset synthesis process is illustrated in Fig.~\ref{fig:data_gen_process}.


\noindent \textbf{Environment Construction}: The simulation environments were modeled using satellite imagery and hand-tuned measurements of the real locations of the corresponding original datasets. During the design phase, special care was taken to replicate the structure of the road, intersection, and background geometry. The precision of the geometry ensures that the resulting point cloud mirrors the spatial distribution seen in the corresponding real-world captures. 

\noindent \textbf{Sensor Modeling}: In the simulation, the virtual lidar sensors were configured to match the real sensor specifications from the target datasets. These specifications include the number of channels, angular resolution, field of view, frame rate, point rate, and measurement range. Sensor placement parameters, including lateral offset from center, height, and tilt, were also aligned with real deployment configurations, enabling a fine match in point density distributions.


\noindent \textbf{Dynamic Content and Annotation}: The dynamic elements in our datasets are generated stochastically. While they do not reproduce the exact historical trajectories (trajectory replay) of the real-world actors (e.g., cars, trucks, cyclists, etc.), their behavior is constrained by the real-world lane topology, traffic signal phases, and physics-based interactions modeled in our custom CARLA maps. This stochastic approach serves as a form of data augmentation, generating unique interaction scenarios that statistically conform to the real site while providing scene diversity. Each generated point cloud frame is annotated with 3D bounding-boxes, object IDs, and semantic tags supporting tasks such as object detection, tracking, and scene understanding.

\begin{table*}[h]
  \caption{An overview of real and digital-twin synthesized replicas of LUMPI \cite{busch2022lumpi}, V2X-Real \cite{xiang2024v2x}, and TUMTraf-I \cite{zimmer2023tumtraf}.}
  \label{tab:dataset_overview}
  \centering
  \begin{tabularx}{\textwidth}{lccccccX}
    \toprule
    \textbf{Dataset} & \textbf{Frames} & \textbf{No.\ of Classes} & \textbf{Object Detection} & \textbf{Tracking} & \textbf{Semantic Segmentation} & \textbf{Instance Segmentation} \\
    \midrule
    LUMPI \cite{busch2022lumpi}        & 90K\textsuperscript{*} & 7  & $\checkmark$ & $\checkmark$ & $\times$ & $\checkmark$ \\
    V2X-Real-IC \cite{xiang2024v2x}   & 4.2K                  & 10 & $\checkmark$ & $\checkmark$ & $\times$ & $\times$    \\
    TUMTraf-I \cite{zimmer2023tumtraf}& 4.8K                  & 10 & $\checkmark$ & $\checkmark$ & $\times$ & $\times$    \\
    \midrule
    \texttt{UT-LUMPI}                  & 10K                  & 6  & $\checkmark$ & $\checkmark$ & $\checkmark$ & $\checkmark$ \\
    \texttt{UT-V2X-Real}               & 10K                  & 6  & $\checkmark$ & $\checkmark$ & $\checkmark$ & $\checkmark$ \\
    \texttt{UT-TUMTraf-I}              & 10K                  & 6  & $\checkmark$ & $\checkmark$ & $\checkmark$ & $\checkmark$ \\
    \bottomrule
  \end{tabularx}
  \begin{flushleft}
    \textsuperscript{*} including both auto-annotated and human-supervised labels.
  \end{flushleft}
\end{table*}

\subsection{Dataset Description}

The synthetic datasets contain 6 classes: Car, Van, Bicycle, Motorcycle, Bus, and Truck. Due to the challenging conformation of pedestrian models in CARLA to real data, all pedestrian and VRU classes are omitted for now to maintain high fidelity, and are planned to be added in a future.

Each synthetic dataset contains 10K frames of intersection focused activity, covering roughly the same area as the target datasets. The datasets include point cloud data along with rich annotations. An overview of the datasets is presented in Table \ref{tab:dataset_overview}.

\section{Dataset Validation and Utility}
\begin{figure*}[h] \centerline{\includegraphics[width=0.98\textwidth]{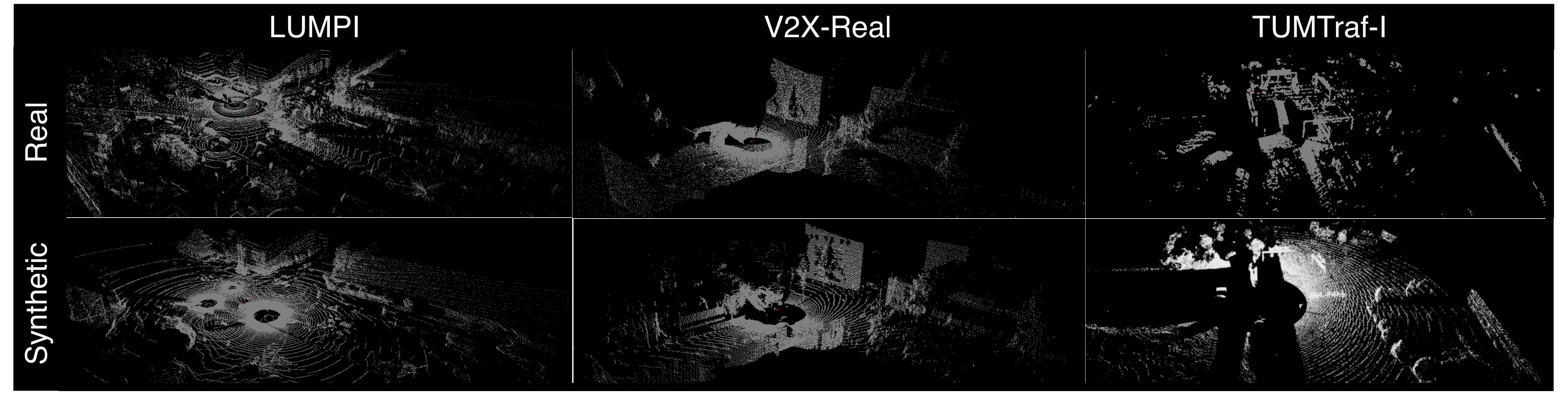}} \caption{A Qualitative Comparison of Real vs. Synthetic Data. A visual resemblance can be noticed in point clouds generated through digital-twin based simulations, to the real point clouds gathered via real lidar sensors. \label{fig:qualitative_comparison}}
\end{figure*}

In order to assess the fidelity and usefulness of the proposed synthetic datasets, this section details a series of experiments and comparative analyses against their real-world counterparts. The objective is to demonstrate that the synthetic datasets exhibit strong structural and distributional alignment, making them well-suited for training, domain transfer, and data augmentation in roadside lidar-based perception pipelines for intelligent transportation systems (ITS). A qualitative comparison of point clouds from the real and synthetic datasets is shown in Fig.~\ref{fig:qualitative_comparison}.

\subsection{Structural and Distributional Similarity}
We begin by evaluating the structural similarity between the synthetic and real datasets using a set of quantitative metrics. These metrics assess distributional alignment within a unified  $80 \times 80 \times 10$ meter\textsuperscript{3}  region, approximately centered around the intersection area, to ensure a fair comparison across datasets. The total number of frames used for comparison varies according to the size of each real dataset. For LUMPI, \textit{Measurement4} is used containing 8120 frames. For V2X-Real, data from an infrastructural lidar from the most completed train set \textit{V2X-Real-Lidar-Cameras} is used, that contains 4169 frames. And for TUMTraf-I, the contiguous subset \textit{R2} sequence \textit{03} is used containing 1033 frames.

\begin{figure*}[h]
\centerline{\includegraphics[width=0.98\textwidth]{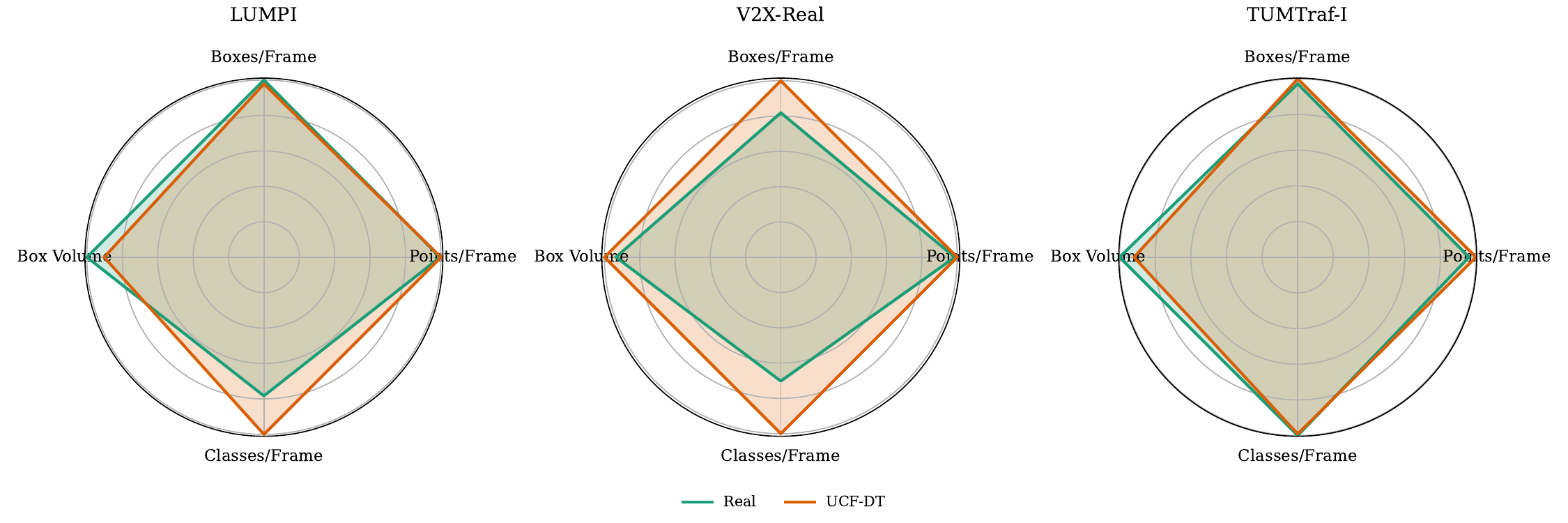}}
\caption{Normalized Frame-Level Means for 4 Key Metrics. Left: LUMPI Dataset, Middle: V2X-Real, Right: TUMTraf-I. The synthetic datasets are carefully generated to match points per frame and boxes (objects) per frame. However, to make the class distribution more even but still comparable to the original datasets' classes, synthetic datasets contain more categories of objects per frame. \label{fig:normalized_frame_level_means}}
\end{figure*}

The comparison is based on metrics that capture scene complexity (average number of objects per frame), point density, object size (bounding box volume), and object class distribution. A summary of these statistics is presented in Fig.~\ref{fig:normalized_frame_level_means}, which shows normalized frame-level means for all of these metrics. A high degree of overlap between the synthetic and real dataset plots indicates that the synthetic datasets closely match the real ones in terms of spatial, object, and class-level distributions. To further assess the alignment between synthetic and real datasets, we analyze each dataset individually, comparing their spatial structure, object composition, and statistical properties.
\vspace{1mm}

\begin{figure}[h]
\centerline{\includegraphics[width=3.5in]{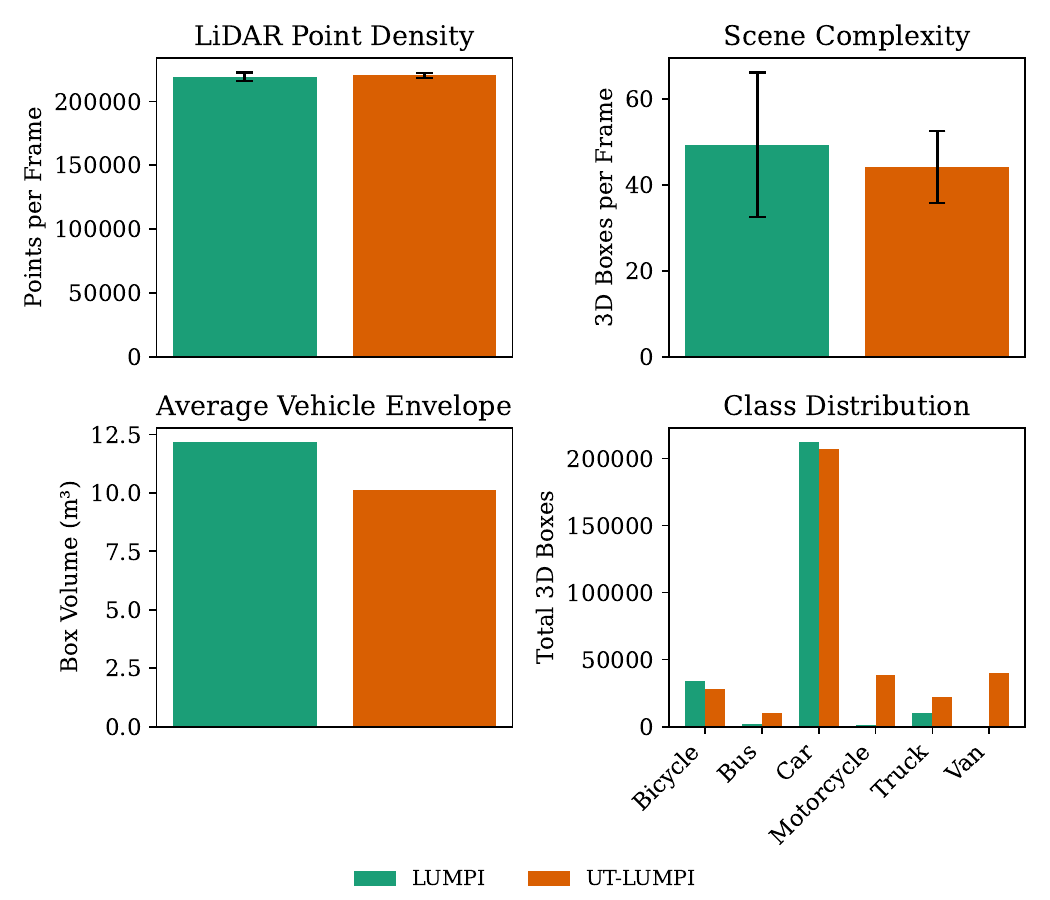}}
\caption{LUMPI vs. \texttt{UT-LUMPI}: Comparison via point density, scene complexity, bounding-box size, and class distribution metrics.
\label{fig:lumpi_composite_plot}}
\end{figure}

\noindent\textbf{LUMPI Dataset:} The LUMPI dataset is a large-scale roadside lidar dataset captured from  Königsworther Platz intersection, located in Hannover, Germany. It contains six classes, that are, \textit{person}, \textit{car}, \textit{bicycle}, \textit{motorcycle}, \textit{bus}, and \textit{truck}. An \textit{unknown} class is also added for non-background miscellaneous objects. The synthetic counterpart, \texttt{UT-LUMPI} is constructed to mimic LUMPI’s spatial layout and sensor specifications. As shown in the block comparison in Fig. \ref{fig:lumpi_composite_plot}, the \texttt{UT-LUMPI} dataset replicates the original's structure closely.

On average, 49.29 objects are present in the original dataset in an $80 \times 80 \times 10$ meter\textsuperscript{3} region centered at intersection area. This is closely matched in the synthetic replica where that average is 44.17. Similarly, mean points per frame in synthetic data equate 220,297.02, matching closely to the average 219,265.08 points in real data. The mean bounding-box volume (10.13 meter\textsuperscript{3}) in simulation dataset also resembles to human-annotated boxes (12.17 meter\textsuperscript{3}). The histogram in the bottom-right panel presents a more even class distribution for \textit{car} and \textit{bicycle} classes. The under represented classes include \textit{motorcycle}, \textit{van}, \textit{bus}, and \textit{truck} where the frequency is increased intentionally to create a more generalized dataset. Though the synthetic data does not contain \textit{person} class, the simulation pipeline supports inclusion of additional classes in future extensions.
\vspace{1mm}

\begin{figure}[h]
\centerline{\includegraphics[width=3.5in]{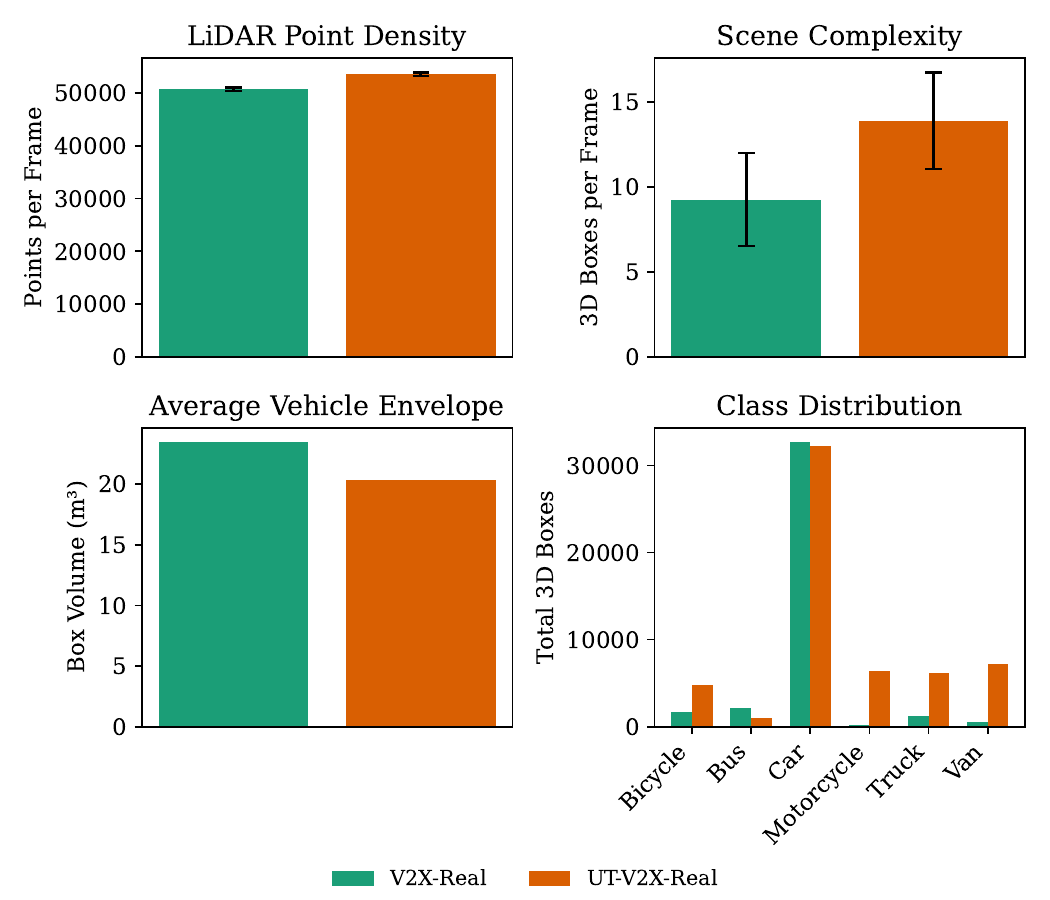}}
\caption{V2X-Real vs. \texttt{UT-V2X-Real}: Comparison via point density, scene complexity, bounding-box size, and class distribution metrics. 
\label{fig:v2xreal_composite_plot}}
\end{figure}

\noindent\textbf{V2X-Real Dataset:}
Originally, the V2X-Real dataset emphasizes connected vehicle scenarios; however, for the sake of this article's focus, only roadside lidar from the south-east corner of Westwood Plaza x Charles E. Young Dr South intersection is considered, which is located within the UCLA campus in Los Angeles, California. This subset is part of the V2X-Real Infrastructure Centric (IC) dataset. It contains 10 classes: \textit{pedestrian}, \textit{scooter}, \textit{motorcycle}, \textit{bicycle}, \textit{truck}, \textit{van}, \textit{barrier}, \textit{box truck}, and \textit{bus}. To keep our synthetic dataset consistent with our other synthetic replicas, the number of classes is kept the same, while ensuring that the synthetic data still matches the real data distributionally. The block metrics shown in Fig. \ref{fig:v2xreal_composite_plot} confirm a strong structural match between V2X-Real and \texttt{UT-V2X-Real}.

The mean number of points per frame in the synthetic \texttt{UT-V2X-Real} dataset is 53,560.59, closely matching the 50,666.87 points per frame in the real dataset. There fewer objects per frame on average (9.24) in real dataset, reflecting the relatively smaller intersection size compared to LUMPI. In contrast, the synthetic dataset averages 13.88 objects per frame, a figure increased to better represent underrepresented classes. The mean bounding-box volumes are, respectively, 21.80 and 26.52 in the synthetic and real datasets. While \texttt{UT-V2X-Real} includes only six object classes, compared to ten in the real dataset, it achieves a more balanced object distribution across classes. As in \texttt{UT-LUMPI}, pedestrian subtypes are excluded from the simulation to avoid domain inconsistencies caused by CARLA’s current limitations in human modeling.

\begin{figure}
\centerline{\includegraphics[width=3.5in]{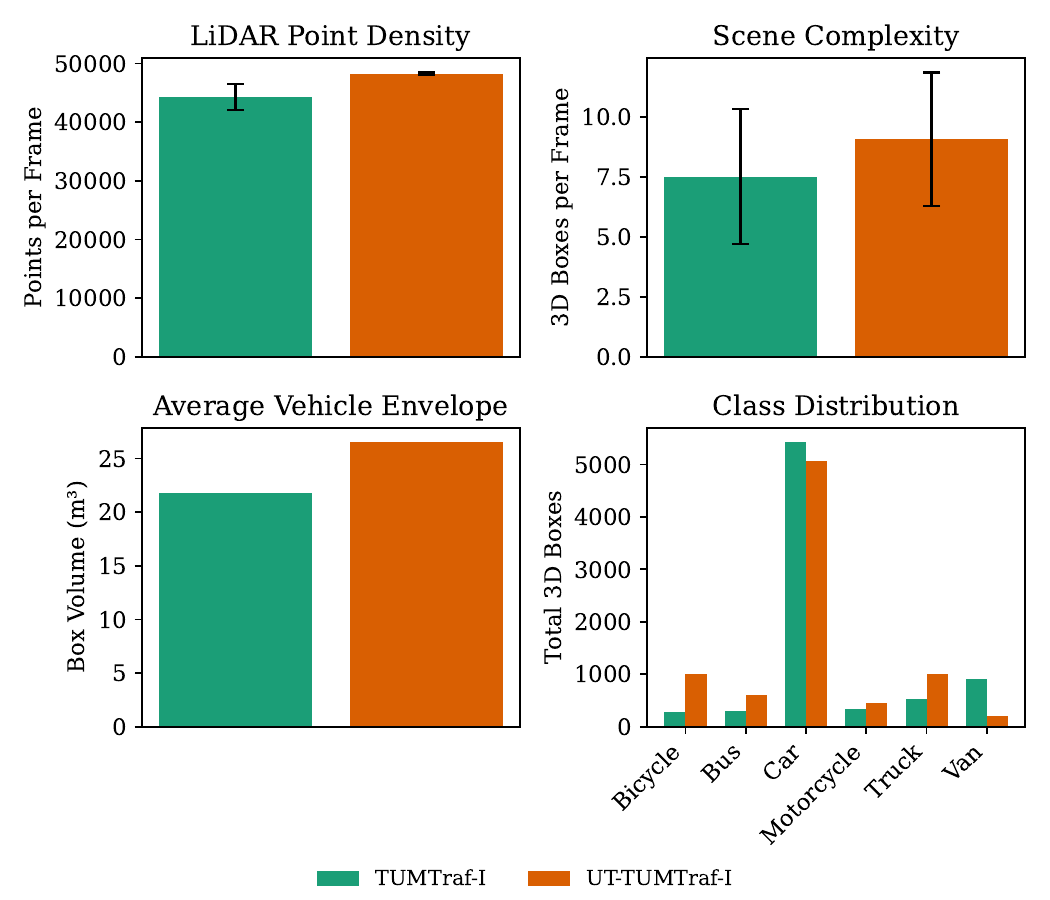}}
\caption{TUMTraf-I vs \texttt{UT-TUMTraf-I}: Comparison via point density, scene complexity, bounding-box size, and class distribution metrics. 
\label{fig:tumtrafi_composite_plot}}
\end{figure}

\noindent\textbf{TUMTraf-I Dataset:}
TUMTraf-I is a more compact dataset focused on a single intersection with moderate traffic activity. Like V2X-Real, it consists of multiple subsets. In this article, we focus on the R2 sequence 03 subset, which provides labeled intersection lidar data. The corresponding synthetic dataset, \texttt{UT-TUMTraf-I}, is constructed to replicate the physical layout of the target location—the Garching bei München intersection in Germany. The simulated sensors are also configured to match the specifications of the real sensors used in the original dataset. Structural similarity between the real and synthetic versions is illustrated in Fig.~\ref{fig:tumtrafi_composite_plot}, where the metrics show a close match across all frame-level dimensions.

The synthetic dataset contains an average of 48,282.18 points per frame, closely aligning with the 44,312.94 points per frame observed in the original dataset. While the original dataset includes an average of 7.5 objects per frame, this number was slightly increased to 9 objects per frame in the synthetic version to better represent previously underrepresented categories. The mean bounding box volume in the synthetic dataset is 26.52 meters\textsuperscript{3}, also comparable to the real dataset’s average of 21.80 meters\textsuperscript{3}.

\begin{figure*}
\centerline{\includegraphics[width=0.98\textwidth]{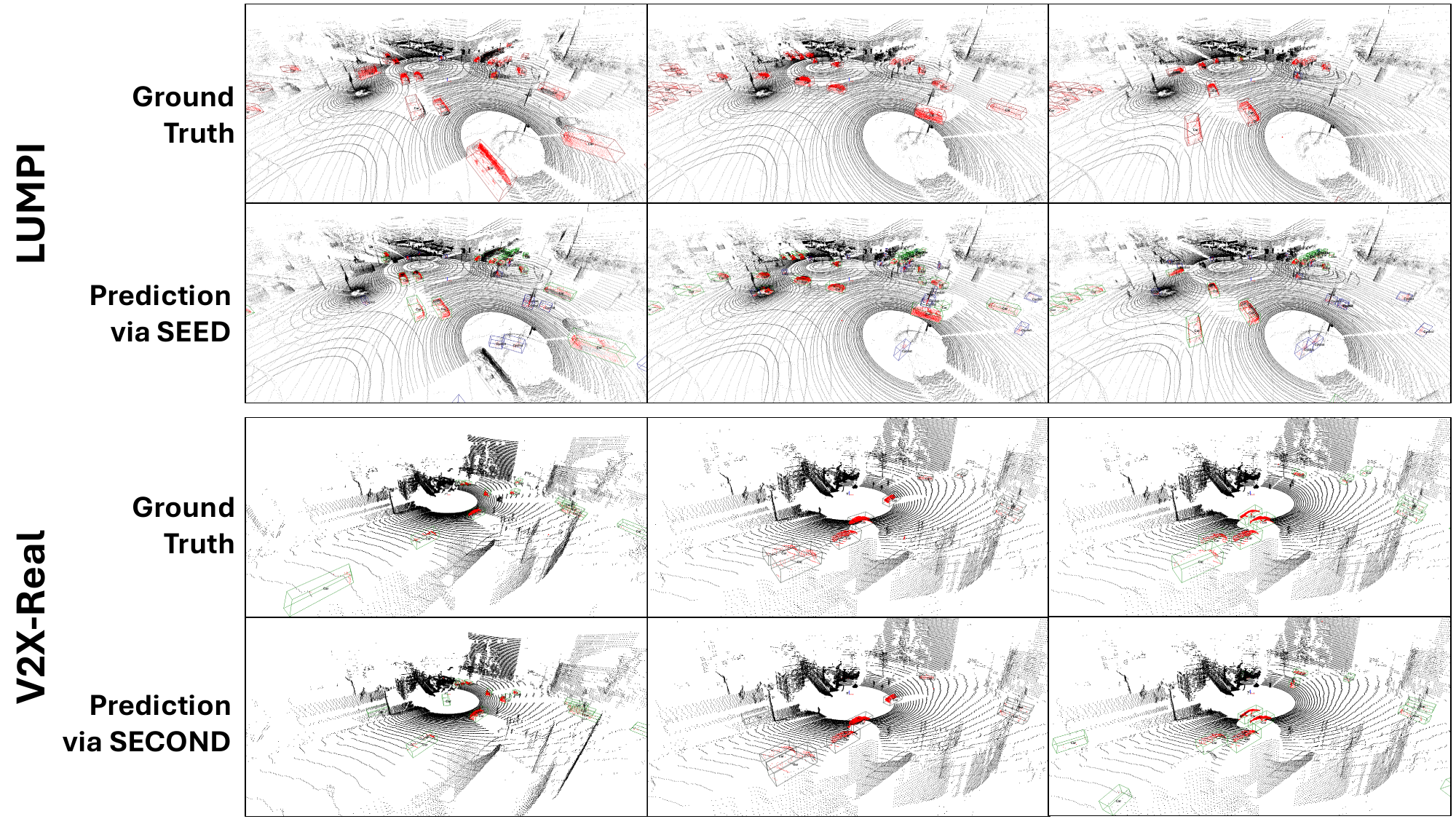}}
\caption{Qualitative Comparison of 3D Object Detection Results on Real Datasets Using Models Trained Exclusively on Synthetic Data. Results from two LiDAR-based object detectors, SEED (top) and SECOND (bottom), trained solely on the synthetic \texttt{UT-LUMPI} and \texttt{UT-V2X-Real} datasets, respectively, and evaluated on their corresponding real-world test sets. Each column shows five randomly selected frames from testsets illustrating ground-truth annotations and model predictions. The close alignment between predicted and ground-truth bounding boxes demonstrates strong cross-domain generalization, confirming that models trained on high-fidelity synthetic datasets can accurately detect real-world objects despite the absence of real training data.}
\label{fig:detection_results_qualitative}
\end{figure*}

\subsection{Utility in Perception Models}
To ensure immediate utility and ease of use, all UrbanTwin datasets have been standardized into a consistent format, facilitating practical application in training and testing 3D perception models. The point clouds are unified (per origin and scale) to target datasets, and are provided as .npy files for easy loading. The labels are formatted to be directly compatible with popular OpenPCDet \cite{openpcdet2020} and SemanticKITTI \cite{behley2019semantickitti} frameworks. Specifically, annotations for 3D object detection (in \texttt{lbl} folder) include object position (\textit{x, y, z}), size (\textit{dx, dy, dz}), and heading (yaw angle around the \textit{z}-axis), with an additional field for the object's tracking ID. For segmentation tasks, labels in \texttt{ins\_seg} and \texttt{sem\_seg} folders (for instance and semantic segmentation respectively), follow the standard SemanticKITTI .label format, enabling research in scene understanding.

All synthetic datasets follow a consistent point cloud and annotation format to ensure ease of use. Owing to their close alignment with real-world counterparts, these synthetic datasets can be used both as high-quality augmentation sources and as standalone datasets for training and evaluation. To complement the structural and statistical analyses presented earlier, we now assess the practical utility of the synthetic datasets through a downstream perception task.

\noindent\textbf{Case Study - 3D Object Detection:}
To demonstrate real-world applicability, we evaluate a common perception task, 3D object detection using the synthetic datasets. The objective of this case study is not to benchmark model performance against state-of-the-art methods or to analyze detection architectures in detail. Rather, it serves as a validation exercise to assess the quality of point clouds, labels, and the overall training utility of the UT-datasets.

To this end, we train two off-the-shelf deep 3D object detectors for the \textit{car} category, SEED \cite{liu2024seed} and SECOND \cite{yan2018second}, on the synthetic \texttt{UT-LUMPI} and \texttt{UT-V2X-Real} datasets, respectively, and evaluate their performance on the corresponding real test sets from LUMPI and V2X-Real. To qualitatively assess this transferability, Fig.~\ref{fig:detection_results_qualitative} provides a visual comparison between ground-truth annotations and the predictions from models trained solely on our synthetic data.

The experimentation is divided into two parts. 1) In the first setup, SEED is trained on the 80\% training split of the real LUMPI dataset and evaluated on the remaining 20\% test set. An identical SEED model—with the same architecture, hyperparameters, and training procedure—is then trained from scratch on the training set of the synthetic \texttt{UT-LUMPI} dataset and evaluated on the same real LUMPI test set. This controlled setup ensures that any observed performance differences can be attributed to the training dataset rather than model-specific factors. 2) In the second setup, SECOND is trained for 56 epochs using the 80\% training split (8,000 frames) of the synthetic \texttt{UT-V2X-Real} dataset. The trained model is then evaluated on the official V2X-Real-IC test set and compared against benchmark results reported in the original V2X-Real paper.

This evaluation is designed to demonstrate the effectiveness of our synthetic datasets in real-world benchmark settings, even when no real training data is used. Importantly, the goal is not to analyze why the trained models may outperform existing baselines, but rather to provide empirical validation of the synthetic datasets’ quality and their ability to generalize to real-world data.

\begin{figure}
\centerline{\includegraphics[width=0.98\linewidth]{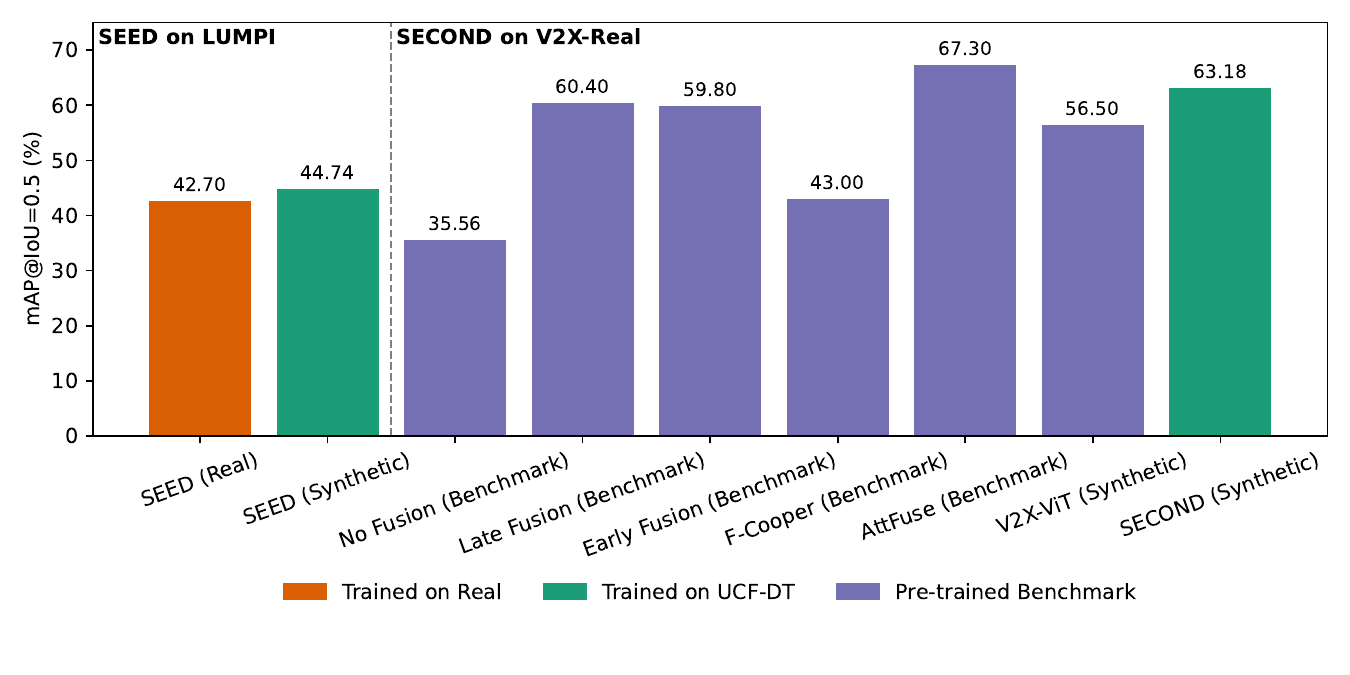}}
    \caption{Comparison of 3D Object Detection Performance for \textit{Car} Category on Real Test Sets for Models Trained on Synthetic Datasets. Left: SEED trained on \texttt{UT-LUMPI} outperforms its real-trained counterpart, validating the utility of simulation-generated annotations. Right: SECOND, trained solely on synthetic \texttt{UT-V2X}, surpasses many reported baselines on V2X-Real, highlighting the effectiveness of digital-twin-synthesized replicas for generalization.
\label{fig:object_detection_benchmark}}
\end{figure}

\vspace{2pt}
\noindent\textbf{LUMPI vs. \texttt{UT-LUMPI}:} For this evaluation, the \textit{Measurement4} is used. This subset, containing 8,120 frames, is further split 80/20, with 6,496 frames used for training and 1,624 kept for testing of the models. In the synthetic dataset, a subset of the first 10,000 frames is taken and is split similarly, 80/20. The larger synthetic training set is intentional, highlighting the advantage of simulation: large volumes of labeled data can be generated at negligible cost, enabling more extensive training. In our experiments, SEED is trained using identical architecture and training hyperparameters. The results, evaluated on the same 1,624 real frames, are presented in Fig.~\ref{fig:object_detection_benchmark}.

Surprisingly, the synthetic-trained model outperformed the real-trained counterpart, despite being evaluated on real annotations. This suggests that \texttt{UT-LUMPI} not only captures the structural complexity of the real dataset but may also offer greater consistency in labeling. The improved performance can likely be attributed to the precision of simulation-generated annotations. In simulations, 3D bounding boxes are generated deterministically based on exact geometric information, whereas human-annotated labels may vary due to occlusions, annotator subjectivity, or limitations of labeling tools. While the absolute gain in mAP is modest, the relative improvement highlights the quality and training utility of the synthetic data.

\vspace{2pt}
\noindent\textbf{V2X-Real vs. \texttt{UT-V2X-Real}:} For V2X-Real, we evaluate the generalization capacity of the synthetic dataset by training the SECOND model exclusively on 8,000 synthetic samples from \texttt{UT-V2X}. The trained model is then evaluated on the V2X-Real-IC test set and compared against benchmark results reported in the original V2X-Real paper. This benchmark includes a diverse set of architectures, ranging from single-stage detectors to more advanced attention and transformer-based models like AttFuse and V2X-ViT, providing a robust test of our dataset's generalization capabilities.

The model achieves Average Precision (AP) of 63.18@3D IoU=0.5 for the \textit{car} class, surpassing most of the previously reported benchmarks in the original V2X-Real paper, even though it was trained without any real data. These results further validate the potential of \texttt{UT-V2X} as a training source, capable of supporting high-performance detection in real-world deployment.

\vspace{2pt}
\noindent\textbf{Scenario Augmentation and Future Support: }The modular design of the simulation pipeline allows for easy scenario augmentation, such as rare-object injection, weather variation, and motion perturbations. While the current release omits pedestrian-related classes due to known limitations in CARLA’s ability to model human behavior accurately, the architecture is fully compatible with their future inclusion once more realistic pedestrian models become available.

\section{Implications on Research and Practice}
The \texttt{UrbanTwin} datasets have significant implications for both research and real-world deployment in the domain of roadside lidar perception. 

\vspace{1mm}
\noindent \textbf{Reducing Cost and Human Effort in Dataset Creation}: From a research perspective, they address a critical limitation: the scarcity and high cost of annotated real-world lidar data. For instance, Amazon SageMaker Ground Truth services cost approximately \$15,000 for labeling 10,000 point cloud frames \cite{sagemaker}—a relatively modest volume by today’s standards. By comparison, the KITTI object detection benchmark, one of the earliest and smallest, contains around 15,000 frames split between training and testing. Modern roadside lidar datasets are much larger: V2X-Real comprises approximately 33,000 lidar frames, while LUMPI contains nearly 90,000 frames with a mix of semi-automated and fully-supervised annotations.

Creating these large-scale datasets demands not only substantial financial investment but also significant human effort and time. For example, the authors of LUMPI report spending 555 hours of expert annotation time to complete their dataset. The scale, cost, and duration required to construct such benchmarks limit their scalability and hinder the widespread development and reliable deployment of lidar-based perception systems.

By offering high-fidelity synthetic replicas of existing lidar datasets at a fraction of the cost and time, \texttt{UrbanTwin} enables broad experimentation across key perception tasks, including 3D object detection, tracking, semantic segmentation, and instance segmentation. Furthermore, the modular simulation framework allows us to extend the UT datasets in future releases to include rare or hazardous scenarios that are difficult or unsafe to capture in the real world. These augmentations can help researchers develop models that generalize to edge cases, ultimately improving the robustness and reliability of lidar-based perception systems in practice.

\vspace{1mm}
\noindent \textbf{Simulation Fidelity and Cross-Task Utility}: The \texttt{UrbanTwin} datasets represent the first effort to leverage a digital-twin modeling approach for data synthesis, incorporating both geometric and dynamic fidelity to real-world settings. In contrast to prior simulation-based methods—many of which rely on fully synthetic environments and hand-crafted assets lacking realism in structure and behavior—\texttt{UrbanTwin} significantly reduces the sim-to-real gap.

Our experiments on the 3D object detection task demonstrate that off-the-shelf models trained on \texttt{UrbanTwin} datasets can be directly transferred to real-world data. This opens up several practical opportunities, from augmenting existing datasets to improve sample and scene diversity, to enabling training for new, unseen locations.

Moreover, these synthetic datasets are applicable to additional perception tasks. For instance, the \texttt{UT-V2X-Real} dataset can be used to train models for semantic or instance segmentation on real-world data—a task unsupported by the original V2X-Real dataset, which lacks segmentation annotations.

\vspace{1mm}
\noindent \textbf{Accurate Labeling and New Research Opportunities}:
In addition to modeling fidelity and cross-task applicability, another key strength of \texttt{UrbanTwin} lies in the quality and consistency of its labels.

Because the \texttt{UrbanTwin} datasets are simulation-generated, they provide consistently accurate annotations, in contrast to human-supervised labeling processes, which may introduce bias or inconsistency. This significantly reduces the effort required to clean and verify datasets, enabling rapid benchmarking and algorithm development at a scale previously constrained by human resources.

Moreover, the availability of accurately modeled digital twins opens up new avenues for methodological innovation—particularly in sim-to-real transfer learning, domain adaptation, and multi-task learning frameworks that span object detection, segmentation, and tracking.

\section{Conclusion \& Future Work}
This paper introduced three high-fidelity synthetic roadside lidar datasets, \texttt{UT-LUMPI}, \texttt{UT-V2X-Real}, and \texttt{UT-TUMTraf-I}, constructed using a digital-twin-based pipeline to replicate the physical layout and sensor properties of real-world deployments. Unlike generic synthetic datasets, our approach achieves close structural, statistical, and sensory alignment with real data.
Through distributional analysis and object detection case studies, we demonstrated that models trained solely on synthetic data generalize well to real-world benchmarks, in some cases outperforming models trained on human-annotated datasets. These results highlight the utility of \texttt{UrbanTwin} datasets as both augmentation and standalone training sources.
By bridging the sim-to-real gap in roadside lidar perception, \texttt{UrbanTwin} offers a scalable, cost-effective resource for both research and deployment in intelligent transportation systems.

While \texttt{UrbanTwin} achieves strong geometric alignment, we acknowledge limitations to be addressed in future. First, strictly stochastic traffic generation, while excellent for increasing sample diversity, may not fully capture the complex, naturalistic interactions (e.g., yielding behavior or aggressive maneuvering) seen in real conflict zones. Future work will explore integrating data-driven behavioral models to enhance the realism of agent dynamics while maintaining the generative flexibility of the simulation. Second, pedestrians, whose inclusion is critical for safety-related ITS research \cite{zheng2025inspe}, were omitted from this release to ensure high fidelity, as standard simulator assets often lack the geometric realism required for Lidar training. We are currently working to integrate realistic human models using Unreal Engine’s MetaHuman framework \cite{epicgames_metahuman_2025} and naturalistic motion from the AMASS library \cite{AMASSICCV2019} to close this gap. Finally, this release focuses on clear weather conditions to establish a baseline for structural fidelity. While the simulation framework supports environmental changes, accurate modeling of Lidar physics in adverse weather (rain, fog) remains a challenge. Future updates will leverage this Digital Twin framework to introduce validated weather conditions, as well as variations in sensor placement and traffic density, to further their generalizability and applicability. We will also expand validation across tracking, segmentation, and sensor fusion tasks, and continue creating synthetic replicas for additional public lidar datasets as part of the UCF Digital Twin initiative.

\printbibliography

\begin{IEEEbiography}[{\includegraphics[width=1in,height=1.25in,clip,keepaspectratio]{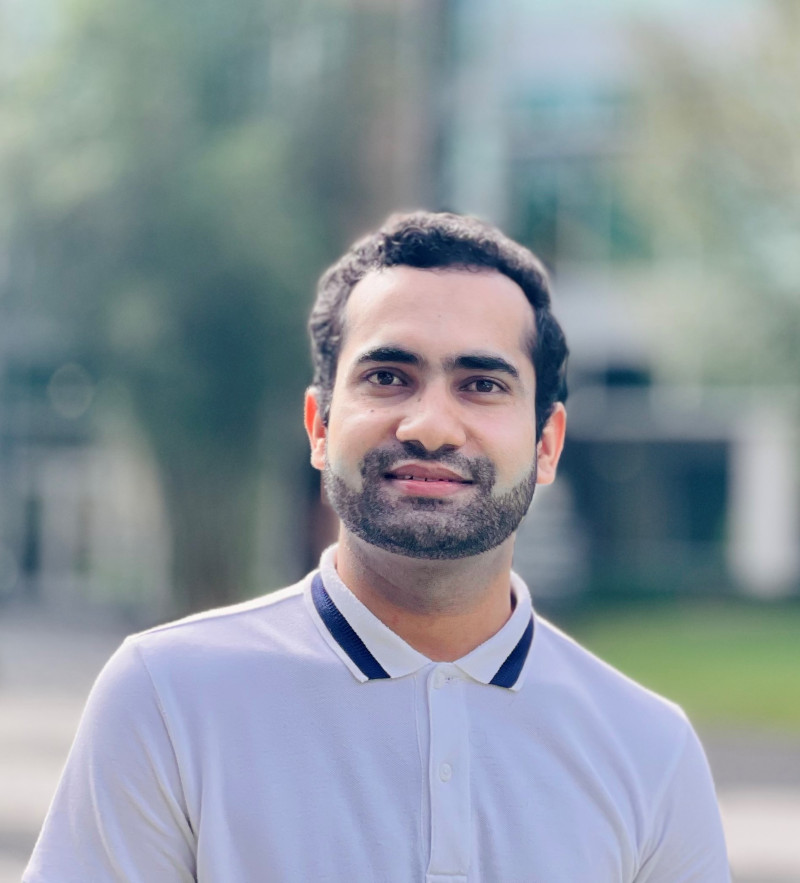}}]{Muhammad Shahbaz } is passionate about interdisciplinary research across Advanced Computer Vision and AI and their applications in the field of intelligent transportation systems and intelligent robotics. He received the B.S. in computer science degree from Pir Mehr Ali Shah Arid Agriculture University Rawalpindi, and M.S. degree in computer science from the Pakistan Institute of Engineering and Applied Sciences, Islamabad, Pakistan. Since 2021, he is pursuing Ph.D. in Civil Engineering at University of Central Florida, USA. His primary focus during Ph.D. spanned efficient and effective 3D perceptions systems for Intelligent Transportation Systems using high-fidelity simulation and multi-modal sensor fusion.
\end{IEEEbiography}

\begin{IEEEbiography}[{\includegraphics[width=1in,height=1.25in,clip,keepaspectratio]{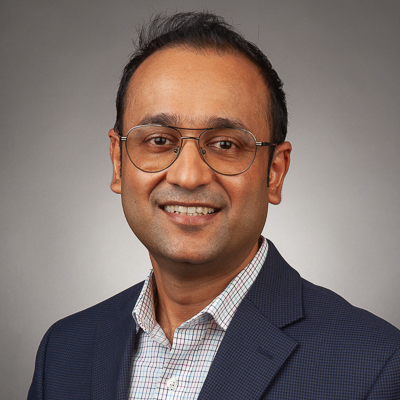}}]
{SHAURYA AGARWAL } 
 (Senior Member, IEEE) is currently an Associate Professor in the Civil, Environmental, and Construction Engineering Department at the University of Central Florida. He is the founding director of the Urban Intelligence and Smart City (URBANITY) Lab, Director of the Future City Initiative at UCF, and serves as the coordinator for Smart Cities Masters program at UCF. He was previously (2016-18) an Assistant Professor in the Electrical and Computer Engineering Department at California State University, Los Angeles. He completed his post-doctoral research at New York University (2016) and his Ph.D. in Electrical Engineering from the University of Nevada, Las Vegas (2015). His B.Tech. degree is in ECE from the Indian Institute of Technology (IIT), Guwahati. His research focuses on interdisciplinary areas of cyber-physical systems, smart and connected transportation, and connected and autonomous vehicles. Passionate about cross-disciplinary research, he integrates control theory, information science, data-driven techniques, and mathematical modeling in his work. As of May 2025, he has published a book, over 37 peer-reviewed journal publications, and multiple conference papers. His work has been funded by several private and government agencies. He is a \textit{senior member} of IEEE and serves as an \textit{Associate Editor} of IEEE Transactions on Intelligent Transportation Systems.
\end{IEEEbiography}

\end{document}